\documentclass{article}

\usepackage{microtype}
\usepackage{graphicx}
\usepackage{makecell}
\usepackage{subfigure}
\usepackage{booktabs} 

\usepackage{hyperref}

\usepackage[accepted]{mlsys2024}

\mlsystitlerunning{Enhancing Cluster Resilience: LLM-agent Based Autonomous Intelligent Cluster Diagnosis System and Evaluation Framework}

\begin{document}

\twocolumn[
\mlsystitle{Enhancing Cluster Resilience: LLM-agent Based Autonomous Intelligent Cluster Diagnosis System and Evaluation Framework}




\begin{mlsysauthorlist}
\mlsysauthor{Honghao Shi}{baai}
\mlsysauthor{Longkai Cheng}{baai}
\mlsysauthor{Wenli Wu}{baai}
\mlsysauthor{Yuhang Wang}{baai}
\mlsysauthor{Xuan Liu}{baai}
\mlsysauthor{Shaokai Nie}{baai}
\mlsysauthor{Weixv Wang}{baai}
\mlsysauthor{Xuebin Min}{baai}
\mlsysauthor{Chunlei Men}{baai}
\mlsysauthor{Yonghua Lin}{baai}
\end{mlsysauthorlist}

\mlsysaffiliation{baai}{Beijing Academy of Artificial Intelligence, Beijing, China}

\mlsyscorrespondingauthor{Yonghua Lin}{yhlin@baai.ac.cn}

\mlsyskeywords{Autonomous Diagnosis, LLM-agent, Cluster Troubleshooting, Evaluation Benchmark}

\vskip 0.3in

\begin{abstract}
Recent advancements in Large Language Models (LLMs) and related technologies such as Retrieval-Augmented Generation (RAG) and Diagram of Thought (DoT) have enabled the creation of autonomous intelligent systems capable of performing cluster diagnostics and troubleshooting. By integrating these technologies with self-play methodologies, we have developed an LLM-agent system designed to autonomously diagnose and resolve issues within AI clusters. Our innovations include a knowledge base tailored for cluster diagnostics, enhanced LLM algorithms, practical deployment strategies for agents, and a benchmark specifically designed for evaluating LLM capabilities in this domain. Through extensive experimentation across multiple dimensions, we have demonstrated the superiority of our system in addressing the challenges faced in cluster diagnostics, particularly in detecting and rectifying performance issues more efficiently and accurately than traditional methods.
\end{abstract}
]



\printAffiliationsAndNotice{\mlsysEqualContribution} 
\section{INTRODUCTION}
Recent advancements in Large Language Models (LLMs) and complementary technologies such as Retrieval-Augmented Generation (RAG) and Diagram of Thought (DoT) have paved the way for the development of autonomous intelligent systems capable of performing cluster diagnostics and troubleshooting. By integrating these technologies with self-play methodologies, we have created an LLM-agent system designed to autonomously diagnose and resolve issues within AI clusters. Our innovative approach includes the establishment of a specialized knowledge base for cluster diagnostics, the enhancement of LLM algorithms to better suit the demands of the domain, practical deployment strategies for agents within real-world environments, and the development of a benchmark specifically tailored to evaluate LLM capabilities in the context of cluster diagnostics. These components collectively contribute to a robust framework that addresses the complexities inherent in managing AI clusters, particularly in scenarios involving performance degradation or other operational anomalies.\\
Through rigorous experimentation, we have validated the effectiveness of our LLM-agent system across multiple dimensions. Our benchmark, which consists of 150 manually crafted advanced questions, serves as a comprehensive evaluation tool that highlights the performance differences between our enhanced LLM-agent and baseline open-source models. In practical applications, the LLM-agent demonstrates its superior capability to identify and resolve performance issues more efficiently than traditional methods, reducing the troubleshooting time significantly. For instance, in a simulated scenario where one GPU was throttled to a much lower frequency, our system identified and resolved the issue within a matter of minutes, whereas conventional approaches would have taken a senior operations engineer nearly an hour to diagnose and rectify using pre-written automated detection software.\\
Moreover, the LLM-agent's ability to detect and initiate corrective actions even before the performance degradation is noticed by human operators marks a significant advancement in proactive system maintenance. This capability not only mitigates immediate issues but also enhances the overall availability and reliability of the cluster by preemptively addressing potential faults. By leveraging the strengths of RAG and DoT, the LLM-agent can autonomously execute remediation measures, thereby freeing up engineering resources to focus on more complex and value-driven tasks. Our research underscores the transformative potential of combining AI-driven diagnostics with practical deployment strategies, setting the stage for a new era of intelligent cluster management solutions.\\
\section{RELATED WORKS}
\subsection{LLM's Alignment and Enhancement}
In recent years, generative artificial intelligence centered around large language models(LLMs) has seen rapid development, with powerful natural language generating capabilities demonstrated by proprietary models such as the GPT series\cite{achiam2023gpt} and Gemini series\cite{team2023gemini}, as well as open-source models like Llama\cite{dubey2024llama} and Qwen\cite{yang2024qwen2}. \\
There are multiple approaches to enhancing the capabilities of LLMs across different stages such as training, inference, and deployment, as well as in areas like data, algorithms, and computational resources. In light of the achievements of autoregressive models like GPT-2(decoder-only transformers)\cite{radford2019language} and LLaMA(transformer++)\cite{touvron2023llama}, enhancing the quality of the data has become a critical method for improving the efficacy of models during the pre-training process\cite{adler2024nemotron,liu2024coachlm}.\\ 
For modern LLMs, there exists several training or fine-tuning works between pre-training and the deployment. ChatGPT\cite{ouyang2022training} describes this process as Supervised Fine-Tuning (SFT), Reward Modeling (RM), and Reinforcement Learning with Human Feedback (RLHF), while LLaMA3.1\cite{dubey2024llama} integrates these into a continuous process known as "Continue Training." Besides training, LLMs can leverage Retrieval-Augmented Generation (RAG)\cite{lewis2020retrieval} to utilize knowledge from data distributions that were not part of the training set. We can refer to the above content as the alignment and enhancement of LLMs.\\
\subsection{AI-agent based Applications}
After the model parameters have been frozen, it is possible to enhance the inherent capabilities of the model through mechanisms such as chain-of-thought(CoT) reasoning\cite{wei2022chain}, scaling test time\cite{snell2024scaling}, and combining CoT LLM and AI agents\cite{castelfranchi1998modelling} as LLM-agent\cite{park2023generative}. \\
CoT is a prompting technique used to guide LLMs to generate intermediate reasoning steps before arriving at a final conclusion. There are extensions to classic CoT, such as Tree of Thought (ToT)\cite{yao2024tree} for tree-like backtracking, Graph of Thought (GoT)\cite{besta2024graph} for graph-based reasoning, and Diagram of Thought (DoT)\cite{zhang2024diagram} for a propose-critique-summarize approach based on topos theory. \\
The development of CoT and the scaling of test-time are unified, with CoT applications always aiming to maintain optimal results with limited test-time or scaling test-time to achieve extraordinaire results\cite{snell2024scaling}. The CoT series technics are also one of the foundations for building LLM-agents. LLM-agents can leverage LLMs as the processing core while integrating traditional AI-agent capabilities such as memory, planning, and execution, creating semi-autonomous software entities that are highly adaptive and capable\cite{xi2023rise}.
\subsection{Diagnosis and Repair for AI Clusters}
Constructing and utilizing LLM applications typically require hardware infrastructure on a scale costing millions of or more dollars. 
Meta constructed the LLM application core LLaMA 3.1 within 54 days, leveraging a cluster that included 16,000 GPUs\cite{dubey2024llama}, with just the GPU costs amounting to over billion dollars. However, such complex and expensive systems face significant challenges in terms of reliability and availability. During the 54-day training, the Meta cluster experienced 419 unexpected interruptions, averaging one disruption every three hours. At such a frequency of interruptions, the cluster, from the operating system to the AI framework and distributed scheduling software, requires the ability to capture, identify, attribute, and repair exceptions to ensure successful and efficient model training. Microsoft's Superbench\cite{xiong2024superbench} has systematically built a suite of standard test cases to comprehensively assess the availability of clusters.\\
In terms of capture and repair, the Torch\cite{paszke2019pytorch} Elastic solution aims to enable automatic restarts of model training, while works such as FlashCheckpointing in DLRover\cite{wang2023dlrover} focus on reducing the cost of checkpoint saving and loading during the automatic restart process. Building upon automatic restart capabilities, many works at the AI framework level have conducted research and practical implementations to enhance reliability and availability, particularly those featuring highly customized solutions based on Megatron\cite{shoeybi2019megatron}. 
ByteDance's Megascale\cite{jiang2024megascale} and Alibaba's Pai-Megatron\cite{qian2024alibaba} both provide toolkits for cluster diagnostics, which are used to check the health of servers and networks, as well as to perform manual or automated error identification and repair. \\
With the advancement of AI technologies, researchers are beginning to explore the use of AI techniques to address cluster diagnostic issues. Using big data techniques to analyze log files was an typical approach to automating cluster diagnostics\cite{jung2021social}. However, such methods primarily involve static or real-time analysis of files produced by the training process, which limits their attribution capabilities and means they lack intelligent autonomy, relying instead on pre-written execution and planning procedure.
\section{SPECIAL TERMINOLOGIES}
AI computing tasks: refers to programs or processes designed to achieve intelligence, such as training large language models, inference with large language models, world model inference, and LLM-agent inference.\\
AI chips: processors suitable for or dedicated to performing AI computing tasks, such as NVIDIA GPUs, Intel Gaudi AI accelerators, and Google TPUs\cite{jouppi2017datacenter}.\\
AI servers: computers equipped with AI chips that are suitable for or specifically designed to perform AI computing tasks, such as the NVIDIA DGX H100. AI servers often have requirements beyond those of classic servers in terms of stability, availability, cooling, and power consumption.\\
AI cluster: a distributed server cluster composed of two or more AI servers set up to accomplish a single target task, such as Meta's cluster containing 16 thousand GPUs. Additionally, AI servers typically require RDMA or higher bandwidth interconnect protocals, such as InfiniBand RDMA\cite{shanley2003infiniband} and RDMA over Converged Ethernet(RoCE)\cite{guo2016rdma}, and do not usually adopt classic Ethernet protocols.\\
Cluster diagnosis: ensuring that AI computing tasks can run with normal performance on the AI cluster, promptly detecting task failures, identifying the points of failure, clarifying the reasons for failure, repairing the corresponding faults, and ensuring the overall availability of the AI cluster.\\
\section{Methods}
\subsection{Overview}
\begin{figure}[htbp]
  \centering
  \includegraphics[width=0.5\textwidth]{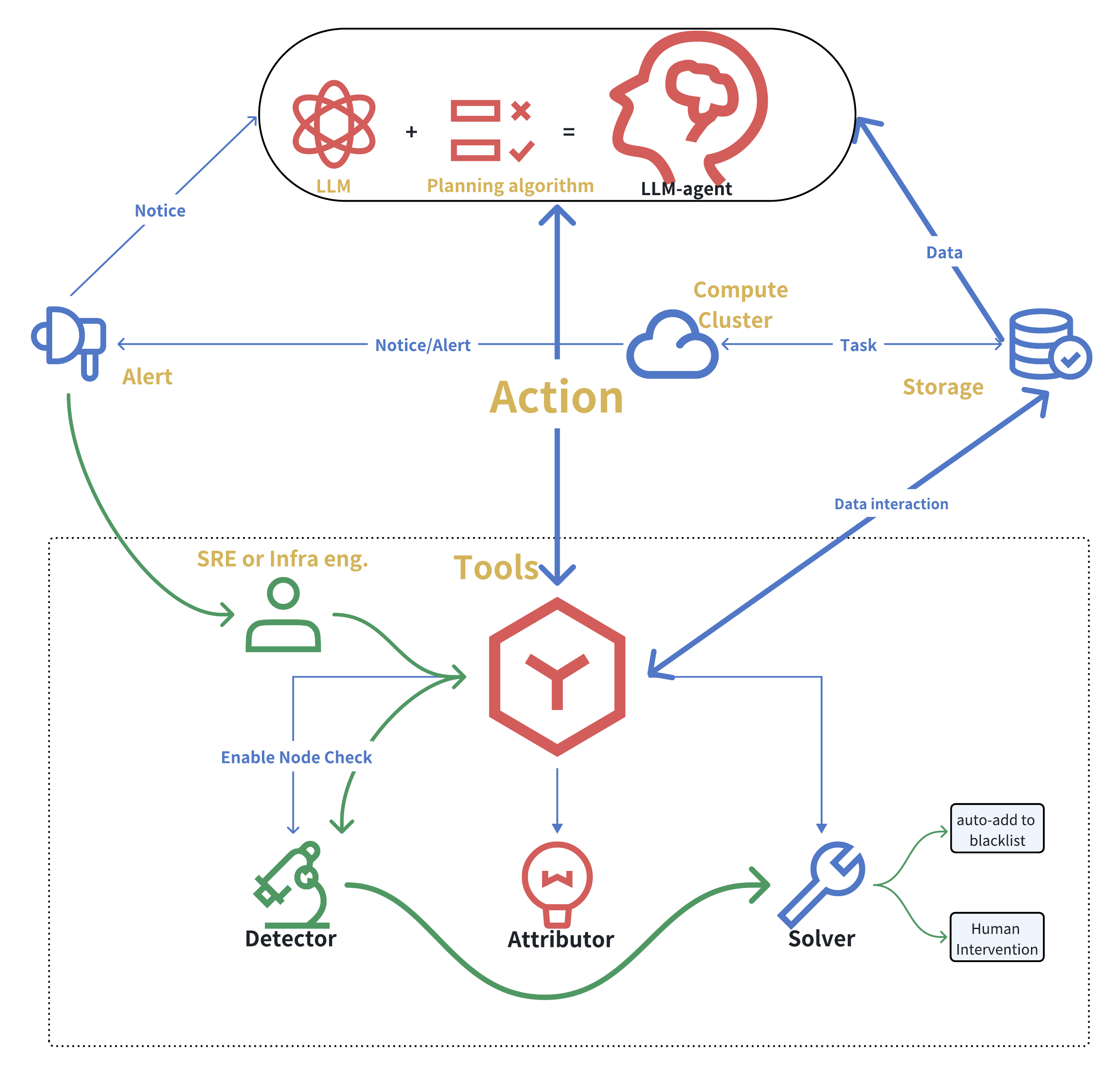}
  \caption{Overview of the Intelligent Maintenance System Based on LLM-Agents
  }
  \label{agent}
\end{figure}

\begin{figure}[htbp]
  \centering
  \includegraphics[width=0.5\textwidth]{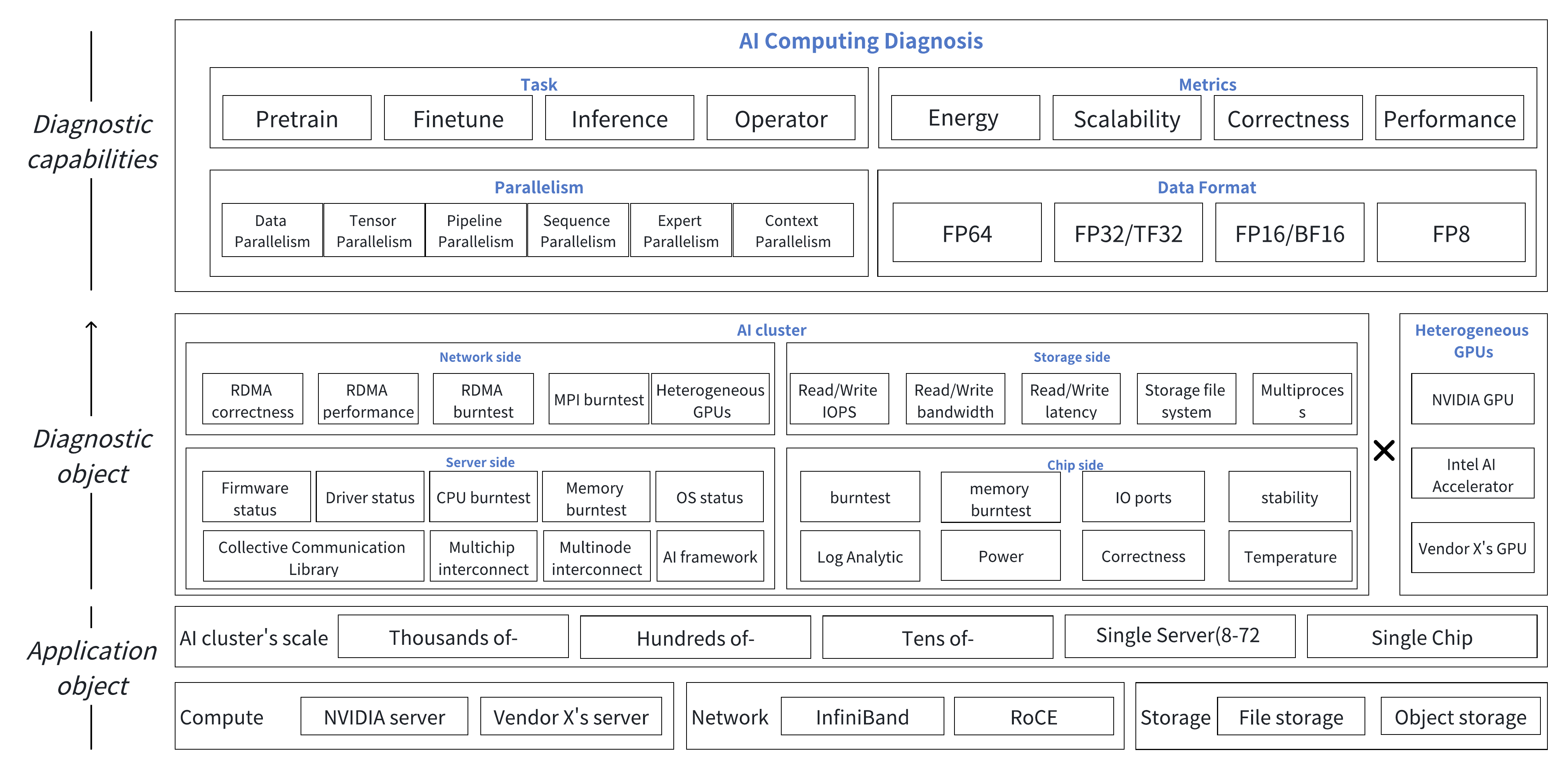}
  \caption{Tools for LLM-agent to Diagnose AI Cluster}
  \label{diagnose}
\end{figure}

We incorporate advanced techniques from the field of LLM alignment and enhancement to creatively develop a solution for building a cluster intelligent maintenance system based on LLM-agents. Figure \ref{agent} illustrates the overall process of this solution. \\
The upper part of the figure represents the core component of solution: the LLM-agent. The LLM-agent consists of an agent program and an LLM. The LLM interprets the input information provided by the agent as external stimuli and task instructions, and responds appropriately. The agent then directly writes code or calls specific software interfaces based on the feedback from the LLM, thereby operating the cluster. For LLM itself, there are two main challenges. First, how does the LLM acquire domain-specific knowledge of cluster diagnostics, and furthermore, where does this knowledge come from. Second, how can the LLM reason and plan? For the entire LLM-agent, ensuring that the LLM's inputs and outputs match with the actual operations performed by the agent controlling the cluster is another crucial aspect that needs to be addressed. \\
In order to solve the above problems, we have introduced three innovations. First, we use 250 cluster failure records collected from GitHub as a starting point, and treat the cluster operation failure logs actually managed by the LLM-agent as a continuous source of data. We utilize RAG\cite{lewis2020retrieval} to enable the LLM to capture detailed knowledge corresponding to specific terms within the context. Figure \ref{agent} describes the "alert", "compute cluster", and "storage sections", along with their communication with the LLM-agent, which outlines this process. Second, we use DoT\cite{zhang2024diagram} enables the model to effectively handle non-natural language information such as symbols, formulas, and code. Similar to vision-text multimodal models, we effectively leverage textual elements that go beyond the inherent meaning of natural language based on DoT. The "planning algorithm" section at the top of Figure \ref{agent} illustrates this innovation. Third, we use self-play technology\cite{snell2024scaling} to enable the LLM to autonomously, also intelligently, devides long tasks or challenging reasoning objectives into multiple steps, self-assess the output of each step, and ultimately achieve the goal. \\
The lower part of Figure \ref{agent} forms the basis of our work. It includes a mature operations alarm troubleshooting and repair process, as well as several mature or advanced software tools. Based on related works, we have developed a unified, multi-level, multi-dimensional cluster diagnostic toolkit as Figure \ref{diagnose}.\\
This tool diagnoses the health status of the cluster from both the supply side and the demand side simultaneously. The bottom part of Figure \ref{diagnose} lists the various components required to build an AI cluster, including the computing component, storage component, network component, and others. AI clusters following different technical routes provide similar capabilities, as shown in the middle part of Figure \ref{diagnose}. We inspect all resource supply items affecting AI computing tasks to determine if their content is correct, if their performance is appropriate, and if they are stable. For example, for the feature of RDMA read/write between two GPUs across servers, our tool checks whether the read/write content is correct, whether the IOPS, bandwidth, latency, and other performance metrics are appropriate, and the stability under complex scenarios such as long-duration or multi-process read/writes. Most of these tools are improved versions of packages provided by chip, server, or operating system vendors. The top part of Figure \ref{diagnose} takes the demand side into consideration, evaluating the metric of concern for AI computing tasks with various characteristics.\\
In summary, we have built an LLM-agent capable of retrieving and utilizing vast amounts of external information, with autonomous planning, learning, reasoning, and execution capabilities. This LLM-agent works alongside either custom-written tools or existing mature tools to perform early warning, troubleshooting, and repair tasks for the cluster.\\
\subsection{Cluster Diagnosis Domain-specific Knowledge Base}
Our knowledge base consists of two sources. One part is logs, monitoring information, or program output content, come from pre-collected, cleaned, and organized GitHub data, carefully selected to address pain points in the cluster diagnostics and troubleshooting domain, incorporating knowledge from issues in the GitHub community, also come from operational data acquired after the initial deployment and operation of the LLM-agent. We call it Diagnosis Dataset. The second part is composed of symbolic reasoning. These reasoning structures use AI computation tasks and hardware specification information as input, and through a bottom-up modeling approach, predict the theoretical performance of the given AI computation tasks, thereby determining the correctness of the performance.\\
\subsubsection{Diagnosis Dataset}
We drew on effective practices from Alibaba’s experience in managing cluster startup operations\cite{xu2024cloudeval} to build a database. We cleaned, organized, and structured the unstructured data obtained from GitHub, ultimately forming an effective dataset. We collected over a thousand questions and feedback items from the GitHub issue section. Through automated processes and manual review, we filtered out over 200 entries with substantive knowledge content and well-structured Q\&A formats. Each piece of organized data contains four fields: problemkey, rawtext, function, and result. \\
The problemkey is a domain keyword identified either manually or based on openai o1. Rawtext refers to the original content of a website after simple formatting, stored as a long string containing the questions asked on the web page and the developers' responses. The function is based on our cluster diagnosis toolkits and is manually correlated by cluster troubleshooting personnel. This part is used as annotation in the portion of the dataset that the model can perceive, it is not perceived by the model for the answers used in the benchmark evaluation part, and it serves as the starting point for knowledge acquisition after the LLM-agent is deployed. The final results are the causes of the faults extracted from the rawtext based on the developers' answers. For an LLM capable of driving an agent to perform cluster diagnostics, we expect it to be able to determine the causes of faults based on real-time operational information from the cluster and to call existing tools or write tool code on-the-fly for cluster repairs, without relying on rawtext containing developer replies. We will demonstrate this capability in subsequent experiments.\\
\subsubsection{Performance Modeling}
We use a series of progressive methods to model the correct performance of given AI computation tasks, and through the DoT, we convert this special modal data into tokens to feed into the model. In addition to cluster health check, we have included modules in the toolkits to determine whether different AI computing tasks exhibit correct performance. These modules can, on one hand, be invoked by the agent to provide results to the LLM for analysis, and on the other hand, they can be called by the LLM to have the agent check the cluster status.\\
We start modeling with the simplest task types. Considering that existing AI clusters are composed of computing devices with the von Neumann architecture, AI computing tasks require the use of computing cores, memory, and I/O ports. It is worth noting that what AI computing tasks occupy are not narrowly defined CPU computing cores, main memory, or input/output ports, but rather in a broader sense, such as computing cores dedicated to matrix multiplication, HBM memory composed of multi-level caches, and high-speed I/O ports formed by PCIe or RDMA protocols. To build a unified model, we use the concepts of equivalent computing power, equivalent memory bandwidth, and equivalent I/O bandwidth.\\
We refer to computational tasks that occupy or primarily occupy one type of resource as single-resource computational tasks. We construct a single-variable computational task performance model and use experiments based on Khinchin's law of large numbers to get the results. We assume that for a certain computational task T, the total amount of resource $R_i$ required is $M_i$. The hardware running this task can provide $N_i$ units of resource $R_i$ per second. Assume that the single-variable task $T_x$ depends only on resource $R_0$. We determine $M_0$ based on the mathematical formula used for the task's computation. For $N_0$, we consider it a random variable. Through a large number of repeated experiments after warm-up, we ensure that the difference between the measured results and the expected value of the random variable approaches zero. We define performance as the number of times a specific task can be executed per unit time. For the aforementioned task $T_x$, we predict its performance to be $\frac{N_0}{M_0}$.\\
For non-single-variable tasks, we focus on modeling whether the different resources they depend on can operate in parallel. A widely used method in multivariate task modeling is the roofline model\cite{ofenbeck2014applying}. The roofline model introduces a new variable: task characteristic $C_T$. The Roofline model introduces a new variable: the task characteristic $C_T$. Consider a task $T_x$depends on two resources $R_0$ and $R_1$, the effective utilization of resource $R_0$ is plotted on the Y-axis, and the ratio of effective utilization of resource $R_0$ to resource $R_1$ is plotted on the X-axis. By changing $C_T$, a scatter plot can be drawn, forming a shape like a roofline. The Roofline model is equivalent to modeling the performance of multivariable tasks under fully parallel scenarios, which does not align with real-world conditions. Additionally, in the context of existing LLM performance modeling, changes in $C_T$ are not about variations in the input size of a single task but about the changing proportions of two different primary resource-consuming tasks within the total task.\\
\begin{figure}[htbp]
  \centering
  \includegraphics[width=0.5\textwidth, trim=0 270 0 50, clip]{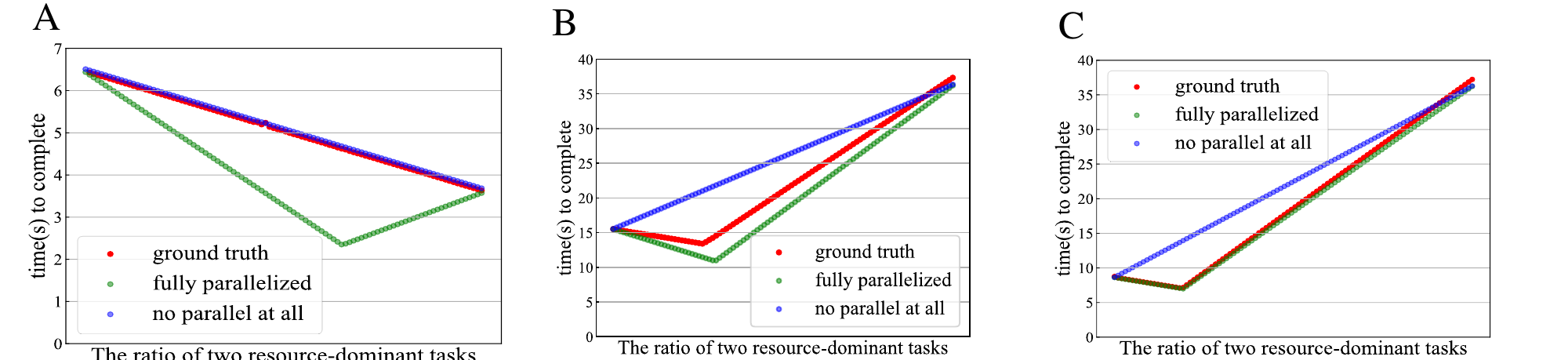}
  \caption{Multi-variable Task Performance Modeling. A shows compute-memory, B shows interconnect-memory, C shows interconnect-compute}
  \label{perf}
\end{figure}
Therefore, we use the proportion of different subtasks as variables to model multivariable tasks for the three main resources provided by AI clusters: equivalent floating-point computing power for matrix multiplication, memory read/write bandwidth, and I/O port bandwidth. The results at figure \ref{perf} show that computing and memory are in domains that are completely non-parallelizable, whereas computing, memory, and I/O ports can approach full parallelization. This conclusion and related figures have been compiled and placed in the RAG documentation.
\subsection{Create LLM-agent with RAG-DoT-Selfplay techniques}
\subsubsection{Using RAG to Build an LLM That Can Utilize External Knowledge}
RAG integrates two core components: retrieval and generation. The retrieval module is responsible for finding context-relevant information from an external knowledge base, a process that typically involves indexing large volumes of documents to quickly locate the most pertinent segments. The retrieved information is then passed to the generation module as additional input. The generation module builds upon a pre-trained language model, leveraging the retrieved context to enhance its generation capabilities, thereby producing responses that are more accurate and better aligned with real-world situations. \\
Considering other similar technologies, SFT requires substantial computing resources and may diminish the model's inherent generalization capabilities. In-context learning consumes context length and inference time, making it unsuitable for importing datasets with millions of entries. RAG can acquire relevant knowledge during inference with minimal resources and inference time, without altering the weights of the model itself.
\subsubsection{Using DoT to Build an Agent That Can Reason and Plan}
DoT(Diagram of Thoughts)\cite{zhang2024diagram} models iterative reasoning in LLMs as constructing a Directed Acyclic Graph (DAG) within a single model. The DAG consists of nodes representing propositions, critiques, refinements, and verifications, with edges indicating the logical relationships or dependencies between them. We use XML to handle multimodal special symbol data and perform reasoning based on DoT.\\
Based on the principles of DoT, we use XML tags to separate different types of text, including plain text, special symbols, code, formulas, and inference rules. Thanks to the rope positional encoding adopted by LLama3.1, the model can accurately capture the content within XML pairs. Based on the reasoning graph, our experiments confirmed that this application allows the LLM to correctly reason according to specific rules, achieving the capability to support the agent in completing cluster fault attribution and repair tasks. This significantly exceeds the capabilities of pre-trained or aligned LLMs.
\subsubsection{Using Selfplay Techniques to Construct a Domain-specific MultiModal Agent}
With the help of RAG and DoT, the LLM can utilize information from outside the training set as well as abstract symbolic reasoning information. However, this still has limitations for an agent designed for intelligent cluster diagnostics. We permit the LLM to generate content over a longer duration. The quality of solutions to challenging problems can be enhanced through multiple rounds of planned selfplay or spontaneous self-questioning and answering by the agent.\\
Spontaneous self-questioning and answering is applied in DoT reasoning. On the planned selfplay process, we transform the complex problem of cluster fault attribution into a three-round process. In the first round, the agent, based on error logs passed from the cluster, prompts the LLM to identify potential keywords from the error items and corresponding solutions from the knowledge base, performing information extraction and RAG. In the second round, the LLM evaluates its own answers, making corrections or accepting them directly, then proceeds to write or call appropriate tools for the Agent to execute. In the final round, the LLM makes an accurate attribution judgment based on the results of the agent's interaction with the actual cluster. Compared to existing selfplay work focused on the text side, we integrate it with the agent, granting it the permissions to operate machines and interact with the environment, fully simulating the capabilities of a human engineer to solve problems.\\
\section{EXPERIMENTS}
We conducted a three-phase experiment to demonstrate the advanced nature of the proposed LLM-agent in the field of cluster intelligent diagnostics. The first phase involves creating a dataset and benchmark for the field of cluster intelligent diagnostics. First, we define the statistical characteristics of the external data knowledge base and introduce the process of generating an evaluation benchmark from this knowledge base. Next, we describe the features of this benchmark and explain its advanced nature in the field of cluster intelligent diagnostics. Throughout this process, we emphasize fairness and impartiality, strictly distinguishing between the parts of the model that can be perceived and the scoring portions of the evaluation. We further elaborate on the benchmark using the results of the mainstream open-source model LLaMA3.1-70B. \\
The second phase involves evaluating the innovative aspects of the three models we proposed—RAG, DoT, and selfplay—using the aforementioned benchmark for comparative assessment. The experiments in the second phase are aimed at demonstrating the advanced nature of our proposed models in the field of cluster intelligent diagnostics. \\
In the third phase, we expose the LLM-agent to both the training and testing sets in the benchmark, allowing it to operate in its most complete form to address real-world problems encountered in production environments. We demonstrate the accuracy, efficiency, and autonomous intelligence of this solution through two typical cases. Specifically, we found that this solution can provide early warnings for AI clusters, further enhancing the availability of the clusters.\\
Finally, we will conduct a qualitative analysis and discussion on the topics of correctness, safety, and reliability, which are at the forefront of the LLM and LLM-agent fields and have yet to be conclusively resolved, to demonstrate the series of work we have undertaken in these areas.\\
\subsection{Statistics and Evaluation for Dataset and Benchmark}
\subsubsection{Data's Source}
The materials provided to the LLM come from three sources. The first source is automatically collected Q\&A data from relevant GitHub communities involved in AI cluster troubleshooting, such as the issue sections of repositories like Megatron, PAI, Deepspeed, and NCCL. This serves as our initial dataset. The data has undergone two rounds of filtering, both automatic and manual, retaining parts with clear solutions and logical dialogues. The second source is the program output obtained by the LLM-agent using RAG+DoT technology on several AI clusters running tasks. These tasks are executed on clusters ranging from 4 to 100 A800 AI servers. The third part consists of special modal data such as symbolic representations and formulas processed using XML according to DoT logic, all of which are unified into the text modality. \\
The total amount of pure text material is 200+ items compared with 1.2GB origin files. This also confirms that if more than 200 items consist of pure text content is fully pre-tokenized to serve as the context for LLM inference, it not only poses a significant challenge to the LLM's capability to handle long texts but also increases the consumption of inference resources, thereby slowing down the execution speed of the LLM-agent.\\
\subsubsection{Benchmark's Source and Statistics for Benchmark}
We divided the original dataset into two parts, approximately in a 20\%-80\% ratio. From the 80\%, we manually compiled 150 questions to assess the LLM's capabilities in the field of cluster diagnostics. During comparative experiments, unless otherwise specified, we provide only 20\% of the original data to all models. During case studies and practical applications, we provide the entire original dataset to the deployed LLM-agent.\\
We designed three evaluation metrics. Metric A evaluates the large model's information extraction capabilities, including extracting the cluster IP addresses and SSH port numbers from conversations, as well as the ability to determine whether further execution is needed, evaluated through string matching. The challenge here is to assess the model's ability to follow instructions and extract information, since logs are derived from user conversations and may contain unnecessary commands that need to be ignored during the determination process. Metric B evaluates the large model's code generation capabilities in the diagnostic domain, including the ability to generate prescribed code based on descriptions given in conversations, control the input and output of the code, and create unseen test cases, implemented in a manner similar to human-eval\cite{chen2021evaluating} but transferred to a real distributed cluster. Metric C evaluates the large model's information attribution capabilities in the diagnostic domain, including the ability to provide attribution based on users' error logs and information. This is currently implemented through multiple-choice questions.
\subsubsection{Evaluation of Benchmark on Standard LLaMA3.1-70B}
We applied this benchmark to several of the most widely used open-source LLMs, namely LLaMA3.1-70B, nemotron-70B\cite{adler2024nemotron}, mistral-120B\cite{jiang2023mistral}, and llama3.2 3B.\\
\begin{table}[t]
\caption{Benchmark's Results on Open-source LLMs}
\label{bench}
\vskip 0.15in
\begin{center}
\begin{small}
\begin{tabular}{
    p{1cm} 
    p{1cm} 
    p{1cm} 
    p{1cm} 
    p{1cm} 
    p{1cm} 
}
\toprule
\makecell[l]{Model} & 
\makecell[l]{Inference\\ on 1 \\A800 GPU} & 
\makecell[l]{Inference\\ in 1 \\A800*8 \\Server} & 
\makecell[l]{Score \\on\\ Metric A} & 
\makecell[l]{Score \\on\\ Metric B} & 
\makecell[l]{Score\\ on\\ Metric C} \\
\midrule
Llama3.1-70B & no & yes & 0.8658 & 0.0 & 0.0 \\
Nemotron-70B & no & yes & 0.7315 & 0.0 & 0.0 \\
Mistral-120B & no & no & 0.7383 & 0.0 & 0.0 \\
Llama3.2-3B & yes & yes & 0.047 & 0.0 & 0.0 \\
\bottomrule
\end{tabular}
\end{small}
\end{center}
\vskip -0.1in
\end{table}
The results is in table \ref{bench}. Due to the lack of relevant data and information, as well as reasoning logic such as DoT, all models were only able to complete the first task, scoring zero on the second and third tasks. Since the results of llama3.2 3B did not meet the minimum requirements for building the LLM-agent, and the 120B model is difficult to infer on a single AI server, we opted for the better-performing and more widely used LLama3.1-70B out of the two 70B models as the basis for subsequent SFT (Supervised Fine-Tuning) and the application of RAG, DoT, and selfplay.\\
\subsection{LMMs' Evaluation}

\subsubsection{Experimental Setup}
We conduct two parts of experiments to comprehensively evaluate and compare the innovative effects of our work. In the first part, we use the mature and universal MMLU\cite{hendrycks2020measuring} benchmark to evaluate the comprehensive ability of the model in basic text understanding after it has been enhanced by RAG, DoT, and self-play. In the second part, through ablation and comparison experiments, combined with the focus areas of the sub-items in our proposed benchmark, we quantitatively demonstrate the advantages of our three innovations.
\subsubsection{General Capability Evaluation Based on MMLU}
Firstly, we aim to substantiate why SFT is not advisable in this domain. Although the LLM that supports the agent needs to possess extensive knowledge in cluster diagnostics, performance modeling, and code writing, we discovered that when the LLM reaches a level where this knowledge can be effectively applied, it often lacks the fundamental interaction capabilities required to engage with the agent. We illustrate this point using the MMLU benchmark.\\
\begin{table}[t]
\caption{MMLU Benchmark's Results on LLama3.1 and Nemotron 70B}
\label{sft}
\vskip 0.15in
\begin{center}
\begin{small}
\begin{tabular}{
    p{2cm} 
    p{2cm} 
    p{2cm} 
}
\toprule
\makecell[l]{Model} & 
\makecell[l]{SFT or not} & 
\makecell[l]{MMLU\\score}  \\
\midrule
Llama3.1-70B & no & 0.8230 \\
Llama3.1-70B & yes & 0.8007 \\
Nemotron-70B & no & 0.8234 \\
Nemotron-70B & yes & 0.7917 \\
\bottomrule
\end{tabular}
\end{small}
\end{center}
\vskip -0.1in
\end{table}
We converted the knowledge repository into tokens compatible with the model and constructed an instruction dataset. We iterated through multiple training rounds until the model could respond correctly to instructions. We then evaluated the SFT model that reached this state against the original open-source model using the Multi-Machine Learning Understanding (MMLU) benchmark. The results are presented in Table \ref{sft}.\\
From the above results, it can be seen that Supervised Fine-Tuning (SFT) leads to a decline in performance when evaluated using general assessment methods such as MMLU. Subsequently, in our proposed cluster diagnostics benchmark, we further observed adverse consequences of this performance decline in metric C. As a result, we ultimately decided not to use the SFT approach to construct the LLM-agent.\\
To avoid the potential risks associated with relying solely on MMLU, we further selected three additional LLM benchmarks that are closely related to the problems we aim to solve in our domain or are entirely generalizable: Abstraction and Reasoning Challenge(ARC)\cite{peter2022abstraction}, BoolQ\cite{clark2019boolq}, and OpenbookQA\cite{mihaylov2018can}. The results are presented in the table \ref{sftall}.
\begin{table}[t]
\caption{Multi Comprehensive Benchmark's Results on LLMs}
\label{sftall}
\vskip 0.15in
\begin{center}
\begin{small}
\begin{tabular}{
    p{1cm} 
    p{0.7cm} 
    p{0.8cm} 
    p{0.8cm} 
    p{0.8cm} 
    p{0.8cm} 
    p{0.8cm} 
}
\toprule
\makecell[l]{Model} & 
\makecell[l]{SFT \\or \\not} & 
\makecell[l]{ARC} & 
\makecell[l]{ARC\\ easy} & 
\makecell[l]{BoolQ} &
\makecell[l]{Open\\bookQA} & 
\makecell[l]{MMLU}  \\
\midrule
Llama3.1-70B & no & 0.6246 & 0.8691 & 0.8786 & 0.3720 & 0.8230 \\
Llama3.1-70B & yes & 0.6032 & 0.8649 & 0.8862 & 0.3680 & 0.8007 \\
Nemotron-70B & no & 0.6280 & 0.8620 & 0.8780 & 0.3680 & 0.8234 \\
Nemotron-70B & yes & 0.6126 & 0.8653 & 0.8859 & 0.3580 & 0.7917 \\
Mistral-120B & no & 0.6544 & 0.8788 & 0.9012 & 0.3980 & 0.8229 \\
Llama3.2-3B & no & 0.4352 & 0.7428 & 0.7835 & 0.2800 & 0.6040 \\
\bottomrule
\end{tabular}
\end{small}
\end{center}
\vskip -0.1in
\end{table}
The results of this set of experiments support the conclusions we drew from the MMLU benchmark.\\
\subsubsection{Results of Our Benchmark}
Table \ref{all} presents all of our experimental results.
\begin{table}[!ht]
\caption{Benchmark's Results on Open-source LLMs(baselines) and our LLM-agent}
\label{all}
\vskip 0.15in
\begin{center}
\begin{small}
\begin{tabular}{
    p{1cm} 
    p{1cm} 
    p{1cm} 
    p{1cm} 
    p{1cm} 
    p{1cm} 
}
\toprule
\makecell[l]{Model} & 
\makecell[l]{"cheating"} & 
\makecell[l]{method} & 
\makecell[l]{Score \\on\\ Metric A} & 
\makecell[l]{Score \\on\\ Metric B} & 
\makecell[l]{Score\\ on\\ Metric C} \\
\midrule
Llama3.1-70B & None & None & 0.8658 & 0.0 & 0.0 \\
Llama3.1-70B & Pre-Written Complete Agent Planning Steps(pre-plan) & None & 0.8658 & 0.4615 & 0.6470 \\
Llama3.1-70B & None & SFT & 0.0 & 0.0 & 0.0 \\
Llama3.1-70B & pre-plan & SFT & 0.0 & 0.9230 & 0.0 \\
Llama3.1-70B & None & RAG & 0.8658 & 0.0 & 0.0 \\
Llama3.1-70B & pre-plan & RAG & 0.8658 & 0.4615 & 0.7059 \\
Llama3.1-70B & None & RAG + DoT + self-play & 0.8466 & 0.6153 & 0.6470 \\
Llama3.1-70B & None & RAG + DoT + self-play + SFT & 0.0 & 0.9230 & 0.0 \\
Llama3.1-70B & whole dataset & RAG + DoT + self-play + SFT & 1.0 & 1.0 & 1.0 \\
Llama3.1-70B & pre-plan + whole dataset & RAG + DoT + self-play + SFT & 1.0 & 1.0 & 1.0 \\
Nemotron-70B & None & None & 0.7315 & 0.0 & 0.0 \\
Nemotron-70B & pre-plan & None & 0.7315 & 0.4615 & 0.7059 \\
Mistral-120B & None & None & 0.7383 & 0.0 & 0.0 \\
Mistral-120B & pre-plan & None & 0.7383 & 0.7692 & 0.8235 \\
Llama3.2-3B &None & None & 0.047 & 0.0 & 0.0 \\
Llama3.2-3B & pre-plan & None & 0.047 & 0.2307 & 0.1176 \\
\bottomrule
\end{tabular}
\end{small}
\end{center}
\vskip -0.1in
\end{table}
The second column of the table indicates whether there was "cheating." We define experiments that do not participate fairly in the benchmark as cheating. While this is unfair for the benchmark portion, it is clearly meaningful for our core research objective: to build an LLM-agent system that can autonomously and intelligently perform cluster diagnostics and troubleshooting. When evaluating the benchmark section, the cheating items can be considered as ground truth.\\
These experimental results can illustrate several conclusions. First, we found that a pre-defined plan can help a naive LLM control the agent. However, this plan was specifically written based on the benchmark questions and cannot be used in a production environment. Correspondingly, all experiments utilizing DoT technology and not cheating scored well on metrics B and C for evaluating the agent, although the scores were slightly lower than those achieved with preplanning. This indicates that our proposed knowledge processing approach based on DoT and self-play can be used to control cluster troubleshooting agents. Second, we found that SFT significantly improved the scores on metric B, which focuses on evaluating code writing or the invocation of diagnostic tools. However, as a trade-off, all models that underwent SFT, even with preplanning, were unable to control the agent properly, resulting in poor performance on metric C. Third, we found that the results based on LLama3.1-70B were not significantly different from those of Mistral-120B, which has nearly twice the number of parameters. Twice the number of parameters implies double or more inference costs (considering multi-GPU linearity), making it impractical. On the other hand, the 3B smaller model, even with preplanning in a cheating scenario, is still unable to handle the task of controlling the agent.\\
We proceeded with subsequent experiments and actual deployment using the LLM-agent enhanced with the whole dataset and all of our innovative methods.\\
\subsection{Intelligent Early Warning and Troubleshooting: A Case Study}
To demonstrate the superiority of the LLM-agent system we have built in the context of intelligent cluster diagnostics, we can present a concrete example to illustrate how the system operates and how it is more efficient and accurate compared to traditional methods. In the production environment of AI clusters, abnormal events or interruptions are not the most challenging problems to resolve. Clear information about anomalies or interruptions can effectively guide senior engineers in diagnosing the causes of issues. Current research is also progressively integrating technologies such as automatic restarts and automatic scheduling into the procedures for handling anomalies or interruptions in AI computing tasks. However, once an AI computing task exhibits slow performance, it becomes difficult to quickly identify the problem, and it is even harder to pinpoint the cause of the slowdown.\\
Assume there is an AI training cluster composed of dozens of servers, where one of the servers suddenly experiences a performance drop. This could be due to various reasons, such as increased network latency, memory leaks, high CPU load, or insufficient storage space. Traditionally, administrators or engineers would check the log files of the cluster to manually identify possible issues. This would involve reviewing logs from different nodes, monitoring system metrics, attempting to reproduce the problem, and so on. This method is time-consuming and labor-intensive and may require multiple attempts to pinpoint the root cause. In our system, the LLM-agent automatically gathers relevant log information, performance metrics, and other necessary data from the nodes of the cluster. Leveraging the LLM-agent’s capabilities assessed through the benchmark, the system extracts useful information from the collected data, such as cluster IP addresses, SSH ports, and other critical diagnostic details. Using its diagnostic capabilities in code generation and information attribution, the LLM-agent identifies the root cause of the issue based on the collected data and information. This may include generating new test cases to validate hypotheses. Once the problem is identified, the LLM-agent generates corresponding remediation scripts and requests human review. After approval, the LLM-agent executes the remediation measures in the cluster. Following the execution of remediation measures, the system collects data again to assess the outcome, forming a closed loop of data, algorithm, and hardware to optimize future diagnostic processes.\\
We manually constructed a scenario. This scenario would lead to slow performance in AI model training tasks and has repeatedly occurred in the development environment. We simulated an extreme heat situation with HVAC failure, throttling the frequency of one of the dozens of GPUs to approximately 200 MHz, rather than the 1410 MHz that the A800 GPUs should operate at. Observing the actual logs shows that the speed of this AI computing task decreased to approximately one-third of its normal performance. Our LLM-system initially flagged the slow AI task through power consumption monitoring and performance modeling results, triggering an automatic alert. Following this, through three rounds of self-play, it recommended checking the GPU core frequencies, a suggestion that the agent then dispatched for execution across all GPUs. Based on the execution results, the LLM accurately pinpointed the GPU with the low core frequency that we had specifically altered. The entire troubleshooting process took less than 10 minutes. In contrast, a senior operations engineer would typically need about one hour to correctly identify the problem and then use a pre-written automated detection software tool created by engineers to determine the specific GPU with the low-frequency fault. More importantly, our LLM-agent can identify the fault before algorithm engineers or operations engineers detect the slow-down phenomenon and automatically complete the repair. This achieves resolving the issue before the fault occurs, thereby enhancing the overall availability of the cluster.\\
\subsection{Qualitative Analysis of Correctness, Safety, and Reliability}
Based on the existing research that is not yet fully mature, and in the context of this specific field of study, we provide reasonable definitions for correctness, safety, and reliability. In this study, we define correctness as whether the process and results of the LLM-agent executing tasks are correct. Compared to evaluating the output of the LLM, assessing the correctness of the LLM-agent's actions is more challenging. An apparently incorrect operation process may produce the correct result, whereas seemingly perfect output at the textual level might lead to an erroneous result when executed. Since we focus on the field of cluster diagnostics with the actual output being the execution of procedures by the agent, we do not investigate the potential harmfulness or bias in the textual content generated by the LLM. Instead, we examine the ability of our LLM-agent to avoid performing harmful operations on the cluster when the information fed back to the agent changes, or even when malicious content is inserted by an attacker, such as deleting files, shutting down, overclocking, or modifying critical system configurations. Regarding reliability, we define it as the overall quality of fault handling by the LLM-agent compared to human engineers or expert human engineers. In addition to whether the attribution is correct, we also consider factors such as the time taken to complete typical fault handling, the resources consumed, and the ability to communicate with non-experts.\\
We incorporate the assessment of correctness into the benchmark evaluation. For the potential risks associated with the LLM-agent, we implement a whitelist plus human review approach. Initially, we ensure the safety of the existing toolkit, followed by creating a whitelist for the program interfaces included in the toolkit and conducting human reviews for the LLM-agent's requests to execute self-authored code. Finally, we observed that the LLM-agent can attribute faults with an average of fewer than three test cases across multiple rounds of self-play, which is more efficient than the twelve cases typically required by human experts. However, regarding communication abilities, the LLM-agent currently does not possess such capabilities. The qualitative analysis described above is mainly aimed at reducing the probability of harmful incidents. Quantitative analysis or a comprehensive model still necessitates further advancements in the field of AI safety.\\
\section{CONCLUSION AND DISCUSSION}
\subsection{Work Summary and Further Plan}
Based on our experience and research in the fields of cluster diagnostics, LLM enhancement, and LLM-agent construction, we innovatively proposed a system solution utilizing LLM-agents to autonomously and intelligently perform cluster troubleshooting. In terms of LLM algorithms, we introduced a benchmark consisting of 150 advanced problems manually crafted, demonstrating the performance differences between our constructed LLM-agent and the original open-source LLMs under fair data conditions. In the realm of LLM-agent construction, we innovatively proposed integrating DoT reasoning mathematics and the ability to handle special symbols and formulas into the agent, enabling the LLM to operate machines at the software level and receive feedback. Ultimately, we applied our innovative achievements to cluster diagnostics, exploring the potential in this field, and were pleasantly surprised to find that the LLM-agent systems, despite being in their extremely early stages, are already capable of handling repetitive and low-end tasks, thus freeing industry practitioners to tackle more challenging and valuable problems.\\
In the future, we will continue our work in four aspects. In terms of LLM algorithms, we will expand and upgrade the existing benchmark and build a more comprehensive and valuable metrics system. In the Agent field, we will further unlock the potential of DoT and make self-written code by the LLM gradually become the main execution body, reducing reliance on preset tools. At the system application level, we will form a closed loop of data, algorithm, and hardware, enriching the database with results from actual deployments. Finally, in terms of safety and reliability, we will continue to work with researchers in related fields to ensure and evaluate the safety and reliability of the agents.
\subsection{Shortcomings and Limitations}
Our research still has shortcomings and limitations. In terms of shortcomings, our agent currently relies on a mechanism of human review to ensure safety, depends on pre-written tools for code, and relies on data sourced from GitHub as a starting point. An ideal LLM-agent system should form a self-sustained relationship with the AI cluster, maintaining and evolving itself. \\
In terms of limitations, our work depends on the LLM within the LLM-agent, but smaller models like llama3.2-3B currently cannot support the capabilities of the agent. Therefore, our work can only be applied to data centers or large-scale distributed clusters and cannot be deployed in edge computing or personal computer scenarios. We need to continuously monitor the development of smaller models and explore the possibility of teaching the capabilities of the LLM-agent to smaller models in the form of DoT when appropriate.

\bibliography{example_paper}
\bibliographystyle{mlsys2024}

\appendix
\section{Please add supplemental material as appendix here}
Put anything that you might normally include after the references as an appendix here, {\it not in a separate supplementary file}. Upload your final camera-ready as a single pdf, including all appendices.


\end{document}